\documentclass{article}

\usepackage{PRIMEarxiv}

\usepackage[utf8]{inputenc} 
\usepackage[T1]{fontenc}    
\usepackage{hyperref}       
\usepackage{url}            
\usepackage{booktabs}       
\usepackage{amsfonts}       
\usepackage{nicefrac}       
\usepackage{microtype}      
\usepackage{fancyhdr}       
\usepackage{graphicx}       
\graphicspath{{media/}}     
\usepackage{cite}  

\usepackage{amsmath,amssymb,amsfonts}
\usepackage{algorithm}
\usepackage{multirow}
\usepackage{algpseudocode}

\usepackage{float}
\usepackage{tabularray}
\usepackage{caption}
\usepackage{subcaption}
\usepackage{textcomp}
\usepackage[normalem]{ulem}
\useunder{\uline}{\ul}{}
\usepackage{longtable}
\usepackage{lineno}
\usepackage{lscape}
\usepackage{amsmath,amssymb,amsfonts}
\usepackage{acronym}
\usepackage{algorithm}
\usepackage{graphicx}
\usepackage{textcomp}
\usepackage{multirow} 
\usepackage{subcaption}
\usepackage[table,xcdraw]{xcolor} 
\usepackage{tabularray}
\usepackage{caption}
\usepackage{array}
\usepackage{geometry}
\usepackage{lscape}
\usepackage{longtable}
\usepackage{pdflscape}

\title{Hybrid Firefly Algorithm and Sperm Swarm Optimization Algorithm using Newton-Raphson Method (HFASSON) and its application in CR-VANET
\thanks{\textit{\underline{Citation}}: 
\textbf{Authors. Title. Pages.... DOI:000000/11111.}} 
}

\author{Rehannara Beegum T, Mohd Yamani Idna Idris, Mohamad Nizam Bin Ayub \\
  Department of Computer System and Technology, Faculty of Computer Science and Information Technology \\
  Universiti Malaya \\
  Kuala Lumpur, Malaysia\\ 
  \texttt{\{Mohd Yamani Idna Idris\}yamani@um.edu.my} \\
   \And
  Hisham A Shehadeh \\
   Department of Information Technology \\
  Al-Huson University College, Al-Balqa Applied
University \\
  Al-Huson, Jordan\\
   \AND
   Usman Ali\\Department of Computer and Software Technology  \\
   University of Swat \\
   KPK, Pakistan \\
}
\begin{document}
\maketitle

\begin{abstract}
This paper proposes a new hybrid algorithm, combining FA, SSO, and the N-R method to accelerate convergence towards global optima, named the Hybrid Firefly Algorithm and Sperm Swarm Optimization with Newton-Raphson (HFASSON). The performance of HFASSON is evaluated using 23 benchmark functions from the CEC 2017 suite, tested in 30, 50, and 100 dimensions. A statistical comparison is performed to assess the effectiveness of HFASSON against FA, SSO, HFASSO, and five hybrid algorithms: Water Cycle Moth Flame Optimization (WCMFO), Hybrid Particle Swarm Optimization and Genetic Algorithm (HPSOGA), Hybrid Sperm Swarm Optimization and Gravitational Search Algorithm (HSSOGSA), Grey Wolf and Cuckoo Search Algorithm (GWOCS), and Hybrid Firefly Genetic Algorithm (FAGA). Results from the Friedman rank test show the superior performance of HFASSON. Additionally, HFASSON is applied to Cognitive Radio Vehicular Ad-hoc Networks (CR-VANET), outperforming basic CR-VANET in spectrum utilization. These findings demonstrate HFASSON’s efficiency in wireless network applications.
\end{abstract}

\keywords{Cognitive Radio-Vehicular Ad hoc network (CR-VANET)  \and Firefly Algorithm (FA) \and Metaheuristic optimization \and Newton-Raphson Method \and Sperm Swarm Optimization (SSO).}

\section{Introduction}
Optimization algorithms are extensively employed by researchers to investigate and develop effective methods for solving a diverse range of complex optimization problems, encompassing both discrete and continuous domains. These algorithms are generally categorized as either deterministic or stochastic. Within the stochastic category, metaheuristic optimization techniques are sophisticated approaches designed to discover optimal solutions while avoiding suboptimal local solutions \cite{siddique2015nature} . These techniques utilize high-level strategies that incorporate stochastic elements, enabling effective navigation of the search space to identify the best solutions. By integrating randomness and adaptive mechanisms, metaheuristics offer robust and efficient solutions to challenging optimization problems. The field of metaheuristics is divided into population-based and trajectory-based search methods, combining insights from individual and collective search strategies to enhance optimization performance\cite{beegum2023optimized}.

A well-defined numerical approach integrated with optimization algorithms, provides in-depth knowledge and understanding of essential factors such as convergence rate and stability of the algorithm\cite{Yang2020b}. This combined methodology enables a thorough analysis of the algorithm's performance, providing a detailed understanding of its efficiency in reaching the optimal solution, the speed at which it converges, and its robustness across varying problem instances. These factors are essential to assess the reliability and effectiveness of optimization algorithms in real-world applications. Several numerical and mathematical models have been employed in the optimization process to achieve quick convergence. Among these, methods such as Gauss-Newton, Quasi-Newton, Newton-Raphson, and conjugate gradients are commonly used. The Newton-Raphson method, in particular, utilizes second-order derivatives instead of relying on gradient descent, which facilitates faster convergence. When the Newton-Raphson method (N-R method) produces a better solution, its step size increases, further accelerating convergence towards the global minimum value \cite{Polyak2007}\cite{Yang2020c}.

In recent years, various metaheuristic algorithms, such as Cuckoo Search, Particle Swarm Optimization (PSO), and the Artificial Bee Colony (ABC) algorithm, have been integrated with the N-R method, yielding promising results across a range of optimization problems. These approaches have been widely reported in the literature, demonstrating significant improvements in solution efficiency and accuracy \cite{Abdel-Baset2016}\cite{Oh2021}\cite{LeDinh2013}. Combining the N-R method with Cuckoo Search has shown enhanced computational precision and convergence effectiveness, particularly in solving linear least squares problems \cite{Abdel-Baset2016}.

Further advancements have been made by merging the N-R method with PSO, a well-known metaheuristic algorithm. This combination, referred to as the "Newton-Raphson Particle Swarm Optimization" (NRPSO), has been employed to evaluate the performance of Bearing-Only Target Motion Analysis (BO-TMA). Through this integration, the algorithm was tested under varying conditions of measurement inaccuracies and the number of measurements, considering three targets and multiple maneuvers. The results confirmed that the NRPSO method provides a robust and adaptable framework for solving complex problems in dynamic environments \cite{Oh2021}.

Additionally, the N-R method has been incorporated into the ABC algorithm to address unconstrained optimization problems, such as optimum power flow. This composite approach enhances the problem-solving capabilities of the ABC algorithm, improving both  accuracy of the solution and convergence rates \cite{LeDinh2013}. Another significant development is the integration of the N-R method with the Sperm Swarm Optimization (SSO) algorithm, resulting in the Modified Sperm Swarm Optimization Algorithm (MSSO). This hybrid approach accelerates the search process, helps avoid local optima, and provides more precise solutions, offering a more effective optimization framework \cite{said2023hybrid}.

These advancements highlight the effectiveness of combining the N-R method with various optimization algorithms to enhance their performance. Building on this trend of hybridization, this paper proposes a novel metaheuristic hybrid optimization strategy that combines the Firefly Algorithm (FA) \cite{yang2010a} with the Sperm Swarm Optimization (SSO) algorithm \cite{Shehadeh2018}, referred to as HFASSO. FA is known for its versatility in handling both continuous and discrete optimization problems, which motivates the selection of FA for this study. Furthermore, FA has been successfully applied in areas such as multimodal optimization, scheduling, and image processing. On the other hand, SSO has proven effective in resolving a variety of optimization challenges, particularly in wireless networks, including Wireless Sensor Networks (WSN) \cite{Shehadeh2018a}. HFASSO aims to achieve the global optimal solution by combining the strengths of both FA and SSO to successfully obtain the global optimal value while minimizing the limitations of each individual algorithm. 

The search capacity and convergence speed are expected to improve when HFASSO is implemented with the N-R method. Therefore, the HFASSO utilizing the N-R method is also attempted, and termed as HFASSON. This article also compares the performance of the HFASSON algorithm with five metaheuristic hybrid algorithms, \textit{viz}., Water Cycle Moth Flame Optimization (WCMFO)\cite{Khalilpourazari2019}, Hybrid Particle Swarm Optimization and Genetic Algorithm (HPSOGA)\cite{Duan2013}, Hybrid Sperm Swarm Optimization and Gravitational Search Algorithm (HSSOGSA)\cite{Shehadeh2021}, Grey Wolf and Cuckoo Search Algorithm (GWOCS)\cite{Xu2017}, and Hybrid Firefly Genetic Algorithm (FAGA)\cite{Rahmani2014}. 

In addition, to validate the efficiency of HFASSON in real-world applications, it is further implemented in the spectrum sensing of CR-VANET. The algorithm possesses numerous advantages, such as its ability to manage both unimodal and multimodal problems, its rapid convergence rate, and its capability to locate the global best solution.

To emphasize the novelty and importance of our work, we would like to outline the key contributions of the proposed algorithm as follows:
\begin{itemize} \item Proposed a new hybrid optimization algorithm, namely Hybrid Firefly Algorithm and Sperm Swarm Optimization Algorithm, and extend it to work effectively by hybridized it with the N-R method. \item Efficiently handles multiple local minima compared to the five selected metaheuristic hybrid algorithms. \item Exhibits a better convergence rate in obtaining the global optimal solution than HFASSO. \item Achieves a more controlled exploration phase at the beginning of the optimization process. \item Demonstrates efficiency in spectrum sensing when applied to Cognitive Radio-Vehicular Ad hoc Network (CR-VANET). \end{itemize}

The rest of this paper includes a background study of FA, SSO, and Newton's Method in Section 2. Section 3 discusses the proposed hybrid algorithm HFASSON. Experimental and Statistical results of the proposed work are analyzed in Section 4 and Section 5, respectively. The implementation of the proposed algorithm with the CR-VANET algorithm is detailed in Section 6. A discussion of HFASSON is provided in Section 7 and conclusions are presented in Section 8.
\section{Background and Rudimentary study of FA, SSO and Newton-Raphson Method}
\subsection{Standard FA and SSO algorithms}
A brief overview of standard FA and SSO algorithms is discussed in this section.
\subsubsection{FA}
FA utilizes the flashing behavior observed in fireflies as its foundational principle. The three simple rules  followed by FA are, \\
Rule 1: Fireflies are considered unisex, making them all attracted to one another without a gender constraint. \\
Rule 2: The level of attractiveness in fireflies is directly correlated with their brightness. A firefly with lower brightness will be attracted to a brighter firefly. As the distance increases, both the level of attraction and brightness decrease. If no firefly is brighter, a firefly will move randomly.\\
Rule 3: The brightness of a firefly is influenced by the environment of the objective function.\\
In most optimization problems, the brightness 
\textit{I} at a specific point x is chosen such that $I(x) \propto f(x)$, where $f(x)$ represents the objective function. However, the attractiveness (represented as $  \beta  $) changes according to the proximity between two fireflies $i$ and $j$. The attractiveness should adjust according to the level of absorption as light intensity decreases and is absorbed by the medium over increasing distances from its source.
The light intensity $I(r)$ varies exponentially with the distance r, which is given by Eq.(\ref{one}).
\begin{equation}\label{one}
I(r) = I_0 e^{-\gamma r} - \gamma r
\end{equation}
where $I_0$ = light intensity;
\\$\gamma$ = light absorption coefficient. 
\\Attractiveness, $\beta$ is calculated as Eq.(\ref{two}).
\begin{equation}\label{two}
\beta=\beta_0e^{-\gamma\left(r_{ij}\right)^2}
\end{equation}
where $\beta_0$= attractiveness at $r = 0$. Generally $\gamma r^{m}$ replaces $\gamma r^{2}$ in Eq.(\ref{two}), allowing more flexibility for exponential decay
based on distance. Here, m is a positive exponent, introducing non-linearity and providing greater flexibility in controlling the rate at which the attractiveness
decreases with distance. 
The distance between the fireflies $i$ and $j$ at positions $x_{i}$ and $x_{j}$ is calculated as Eq.(\ref{three}).
\begin{equation}\label{three}
r_{ij} = x_i-x_j
\end{equation}
The movement of a firefly towards another attractive firefly, regulated by $x_i$ in Eq.(\ref{four}), determines the manner in which a firefly moves towards its target.
\begin{equation}\label{four}
x_i = x_i+ \beta \left(x_i-x_j\right)+\alpha\varepsilon_i 
\end{equation}
For most cases during implementation, $\beta_{0}$ = 1, $m=2$, $\alpha$ ranges between 0 and 1 and $\gamma$ as 1. \(\varepsilon_i\) is a vector of random variables sampled from a uniform distribution with values ranging from 0 to 1.
\subsubsection{SSO}
SSO is based on the biological process through which sperm fertilize an egg, drawing on knowledge of reproductive biology. In SSO, sperm advance from the cervix, which has a lower temperature than the surrounding area. They then migrate towards the Fallopian tubes, a warmer environment where fertilization can occur, in search of the egg. This area is considered the optimal destination for the swarm. Knowledge of how vaginal pH and temperature influence sperm migration is crucial, as these factors significantly affect the rate and direction of sperm movement. The mobility of sperm is significantly influenced by these variables. The initial velocity of sperm is obtained by Eq.(5).
\begin{equation} \label{five}
V_0 = d_{f}\times V_{sperm}\times \log_{10}(pH_{rand1})
\end{equation}
Where:
\begin{itemize}
\item $d_f$: an arbitrary value between 0 and 1 representing a velocity dampening factor that modifies the sperm velocity.
\item $V_{sperm}$: sperm velocity
\item $pH_{rand1}$: a random number between 7 and 14 that indicates the pH value. 
\end{itemize}
The current best solution, denoted $Best\_Solution_{Current}$, refers to the optimal solution achieved so far. The sperm's current best solution is represented by Eq.(\ref{six}).
\begin{equation}\label{six}
\begin{split}
\text{Best\_Solution}_{\text{Current}} &= \log_{10}(pH_{\text{rand2}}) \times \log_{10}(Temp_{\text{rand1}}) \\
&\quad \times (Solution_{\text{sperm}}{\text{[]}} - \text{Current[]})
\end{split}
\end{equation}
Where:
\begin{itemize}
\item $Solution_{\text{sperm}}[]$: the best result obtained by the sperm so far.
\item $pH_{\text{rand2}}$: the pH value, a random integer between 7 and 14.
\item $Temp_{\text{rand1}}$: the temperature, ranging between 35.1\textdegree C and 38.5\textdegree C\cite{Shehadeh2018}.
\end{itemize}
The sperm with the most accurate information at the moment is represented by the global best value. The following Eq. \ref{seven} represents the global best value of sperm:
\begin{equation} \label{seven}
\begin{split}
\text{Best\_Solution}_{\text{Global}} &= \log_{10}(pH_{\text{rand3}}) \times \log_{10}(Temp_{\text{rand2}}) \\
&\quad \times (Solution_{\text{Global\_sperm}}[] - \text{Current}[])
\end{split}
\end{equation}
Where:
\begin{itemize}
\item $Solution_{\text{Global\_sperm}}[]$ is the best solution achieved so far.
\item $pH_{\text{rand3}}$ is the pH value, ranging from 7 to 14.
\item $Temp_{\text{rand2}}$ is a random number between 35.1\textdegree C and 38.5\textdegree C, representing the temperature of the area.
\end{itemize}
$\text{Current}[]$ is the current best solution given by the Eq.(\ref{eight}).
\begin{equation} \label{eight}
\text{Current}[] = \text{Current}[] + \text{Velocity}_{\text{Sperm}}[]
\end{equation}
The sperm velocity update is represented by the following equation:
\begin{equation} \label{nine}
\text{Velocity}_{\text{Sperm}}[] = V_0  + \text{Best\_Solution}_{\text{Current}} + \text{Best\_Solution}_{\text{Global}}
\end{equation}
The sperm velocity in Eq.(\ref{nine}) represents three distinct velocities of sperm used to reach the optimal solution. The pH value of the cervix zone has an impact on the initial velocity. The individual current ideal solution of the sperm, noted as its best solution, is referred to as the second velocity. Sperm travel at a certain speed in response to temperature and pH in the visited zone, which aids in locating the egg. 
\subsubsection{Newton-Raphson Method (N-R Method)}
N-R method is employed in numerical analysis to discover roots within functions of real value. The approach generates progressively better estimates for the roots (or zeros) of a real-valued function\cite{Garrett2015}. The algorithm begins with an initial conjecture $x_{0}$, representing an estimate of a potential root. By leveraging the tangent line of $f(x_{0})$ at the initial estimate, the algorithm computes a subsequent approximation, $x_{1}$. This point gives the precise location where the tangent line intersects the x-axis.
 \begin{equation}\label{ten}
     x_{\text{1}} = x_{\text{0}} - \frac{f(x_{0})}{f'(x_{0})}
 \end{equation}
The expression provided by Eq.(\ref{ten}) estimates \( x_{\text{1}} \) based solely on the known characteristics of the function at \( x_{\text{0}} \). This estimation assumes that the derivative of the function, \( f'(x_0) \), is zero at \( x_{\text{0}} \) indicating that \( x_{\text{0}} \) is a critical or stationary point where the function's rate of change is zero. By making this assumption, the expression allows for the approximation of the next point of interest, \( x_{\text{1}} \), based solely on the \text{function's} properties at \( x_{\text{0}} \), without the need for additional information..

Eq.(\ref{ten}) serves as a concise representation of a single iteration within the N-R method. Until a sufficiently precise result is obtained, this iterative procedure is carried out repeatedly. Thus, Eq.(\ref{ten}) can be generalized and represented as Eq.(\ref{eleven}):
\begin{equation}\label{eleven}
       x_{\text{next}} =x_{\text{prev}} -\frac{f(x_{\text{prev}})}{f'(x_{\text{prev}})}
\end{equation}
Through the systematic application of these meticulously designed steps, the N-R method astutely refines each approximation, gradually unveiling the elusive roots with remarkable precision. 
Consider the minimization of some function, \( g(x) \), using Newton’s method. Then the minima of \( g(x) \) corresponds to points where its first derivative, \( g'(x) = 0 \). Thus by considering \( f(x_{\text{prev}}) = g'(x_{\text{prev}}) \) and \( f'(x_{\text{prev}}) = g''(x_{\text{prev}}) \), minimization problems can be optimize by substituting the values of \( f(x_{\text{prev}}) \) and \( f'(x_{\text{prev}}) \) into Newton’s equation, Eq.(\ref{eleven}):
\begin{equation}\label{Newtonoptimization}
x_{\text{next}} = x_{\text{prev}} - \frac{g'(x_{\text{prev}})}{g''(x_{\text{prev}})}
\end{equation}

\section{Hybrid Firefly Algorithm and Sperm Swarm Optimization with Newton's Method (HFASSON)}
In HFASSON, fireflies are employed to exploit the regional optima, while sperm cells are utilized to search the search space. The sperm cells are attracted to the fireflies with higher brightness within the search space, while the fireflies are guided towards the optimal solution identified by the sperm cells.

The detailed steps involved in HFASSON are described in Algorithm \ref{algo}. The algorithm performs better when exploration and exploitation are effectively combined. Furthermore, the technique tends to converge more quickly and effectively when implemented with a numerical approach, such as the N-R method.
\begin{algorithm}
    \caption{HFASSON}
    \label{algo}
    \begin{algorithmic}[1]
        \State Begin
        \State Initialize population size (\(n\)), \(\gamma\), \(\beta_0\), \(\alpha\), \(\delta\), pH value, temperature, Maximum iteration value (\(iter_{\text{max}}\)), $w_{\text{min}}=0.2$, $w_{\text{max}}=0.9$.
        \State Randomly initialize position, velocity of the particles.
        \For{Iteration = 1: \(iter_{\text{max}}\)}
            \For{all particles}
                \State Calculate the fitness value of the agents.
                \State Update the local and global optimum solutions.
            \EndFor
            \State Update Inertia Weight \cite{Dye2010},
             \( w = w_{\text{max}} - \text{Iteration} \times \left( \frac{w_{\text{max}} - w_{\text{min}}}{iter_{\text{max}}} \right) \).
            \State Update the position using Eq(\ref{thirteen}).
            \State Update the velocity using Eq(\ref{twelve}).
            \State Numerically approximate the first and second derivatives of the fitness function using the N-R method.
            \State Update the position based on the above derivatives.
            \If{(End criteria reached?)}
                \State go to Step 21.
            \Else
                \State go to Step 4.
            \EndIf
        \EndFor
        \State Display the optimal solution.
        \State Stop
    \end{algorithmic}
\end{algorithm}

The hybrid algorithm takes advantage of the exploratory characteristics of the Firefly Algorithm (FA) and leverages the exploitation capabilities of the Sperm Swarm Optimization (SSO) algorithm. The rationale behind combining these two algorithms lies in the knowledge that the FA utilizes the concept of brightness or light intensity for exploration, whereas the SSO algorithm employs linear motion for locomotion. 

By integrating both methodologies, the hybrid algorithm maximizes the benefits of each approach, facilitating efficient navigation and optimal utilization of the search space to identify superior solutions. This fusion leverages the unique characteristics of the Firefly Algorithm (FA) and Sperm Swarm Optimization (SSO), resulting in improved performance through a synergistic combination of their distinct capabilities.

The proposed algorithm, HFASSON, is mathematically represented as follows:
Building upon Eq. (\ref{nine}), Eq. (\ref{twelve}) is derived to calculate the velocity of the agents by incorporating the methods of the hybrid algorithm described previously.
\begin{equation} \label{twelve}
\begin{split} 
V_{Agent}(i+1) = w \times \log_{10}(pH_{\text{rand1}}) \times V_{Agent}(i)  \\  + \log_{10}(pH_{\text{rand2}})\times\log_{10}(Temp_{\text{rand1}})\\
\times(Solution\_Global)_{Agent}[] - {current[]})
\end{split}
\end{equation}
Where: 
\begin{itemize}
\item w= inertia weight\cite{Bansal2011}, values ranging from 0 to 1.
\item $(Solution\_Global)_{Agent}[]$, the best solution currently attained by any agent.
\item pH has a range of 7 to 14.
\item $T$ is a randomly generated value in the range of 35.1 to 38.5, representing the ambient temperature.
\item current[] refers to the best solution currently obtained by Eq.(\ref{eight}).
\end{itemize}
The algorithm iteratively updates the positions of the agents based on Eq. (\ref{thirteen}), systematically exploring the solution space in an effort to identify optimal or near-optimal solutions.
\begin{equation} \label{thirteen}
X(i+1) = X(i) + V(i+1)
\end{equation}
Here, 
\begin{itemize}
    \item $X(i)$ is the current position of the sperm during $i\textsuperscript{th}$ iteration.
    \item $X(i+1)$ denotes the updated position of the agent at $(i+1)\textsuperscript{th}$, and
    \item $V(i+1)$ is the velocity or movement vector of the sperm at $(i+1)\textsuperscript{th}$.
\end{itemize} 

HFASSON starts by randomly initializing all agents, where each agent represents a candidate solution. The $V(i)$ of all agents is determined using Eq.(\ref{twelve}) after updating the $(Solution\_Global)_{Agent}$ throughout each iteration of the algorithm. Finally, using Eq.(\ref{thirteen}), the agents' locations are updated, and the location will converge using the N-R method as well. Until an end condition is met, the position and velocity updating procedure will continue. 
\begin{table*}[htbp]
\caption{Objective Function description}
\label{table1}
\resizebox{\textwidth}{!}{%
\begin{tabular}{lll}
\hline
\textbf{Function} & \textbf{Dimension} & \textbf{Range}     \\ 
\hline
$f_{1}\left( y\right) =\sum ^{d}_{i=1}y_{i}^{2}$                &30, 50, 100                 & [-100,   100]       \\
$f_{2}\left( y\right) =\sum ^{y}_{i=1}\left| y_{i}\right| +\prod ^{d}_{i=1}\left| y_{i}\right|$              & 30, 50, 100                 & {[}-100,   100{]}         \\
$f_{3}\left( y\right) =\sum ^{d}_{i=1}\left( \sum ^{i}_{j-1}y_{j}\right) ^{2}$                & 30, 50, 100                 & {[}-100,   100{]}       \\
$f_{4}\left( y\right) =\max _{i}\left\{ \left| y_{i}\right| ,1\leq i\leq d\right\}$               & 30, 50, 100                 & {[}-100,   100{]}       \\
$f_5\left(y\right)=\sum_{i=1}^{d-1}\left[100\left(y_{i+1}-y_i^2\right)^2+\left(y_i-1\right)^2\right]$                & 30, 50, 100                 & {[}-30,   30{]}         \\
$f_6\left(y\right)=\sum_{i=1}^d\left(\lfloor\ y_i+0.5\rfloor\right)^2$                & 30, 50, 100                 & {[}-100,   100{]}       \\
$f_7\left(y\right)=\sum_{i=1}^diy_i^4+random[0,1)$                & 30, 50, 100                 & {[}-1.28,   1.28{]}     \\
$f_{8}\left( y\right) =\sum ^{d}_{i=1}-y_{i}\sin \left( \sqrt{\left| y_{i}\right| }\right)$               & 30, 50, 100                 & {[}-500,   500{]}       \\
$f_9\left(y\right)=\sum_{i=1}^d\left[y_{i}^{2}-10\cos\left(2\pi y_i\right)+10\right]$                & 30, 50, 100                 & {[}-5.12,   5.12{]}     \\
$f_{10}\left(y\right)=-20\exp\left(0.2\sqrt{\frac{i}{y}\sum_{i=1}^dy_i^2}\right) -\exp\left(\frac{i}{y}\sum_{i=1}^d\cos\left(2\pi y_i\right)\right)+20+e $               & 30, 50, 100                 & {[}-32,   32{]}         \\
$f_{11}\left( y\right) =\dfrac{i}{4000}\sum ^{d}_{i=1}y_{i}^{2}-\prod ^{d}_{i=1}\cos \left( \dfrac{y_{i}}{\sqrt{i}}\right) +1 $               & 30, 50, 100                 & {[}-600,   600{]}       \\
$f_{12}\left( y\right) =\dfrac{\pi }{d}\{ 10\sin \left( \pi y_1\right) +\sum ^{d}_{i=1}\left( v_{i}-1\right) ^{2}\left[ 1+10\sin ^{2}\left( \pi v_{i+1}\right) \right] +\left( v_{y}-1\right) ^{2}\} +\sum ^{y}_{i=1} u\left( y_{i},10,100,4\right)$               & 30, 50, 100                 & {[}-50,   50{]} \\ 
$f_{13}(y) = 0.1 \left\{
    \sin^2(3 \pi y_1) + \sum_{i=1}^{z} (y_i - 1)^2 \left[1 + \sin^2(3 \pi y_i + 1)\right]
    + (y_z - 1)^2 \left[1 + \sin^2(2 \pi y_z)\right]
    + \sum_{i=1}^{z} u(y_i, 5, 100, 4) \right\}$    & 30, 50, 100                 & {[}-50,   50{]}
    
\\
$f_{14}(y) = \left(
    \frac{1}{500} + \sum_{j=1}^{25} \frac{1}{j + \sum_{i=1}^{j} (y_i - a_{ij})^6}
\right)^{-1}$
    & 30, 50, 100                   &
    {[}-62.536, 65.536{]} \\
$f_{15}(y) = \sum_{i=1}^{11} \left[
    a_i - \frac{y_1(b_{i1}^2 + b_{i2} y_2)}{b_{i1}^2 + b_{i3} y_3 + b_{i4} y_4}
\right]^2$
    & 30, 50, 100                   &
    {[}-5, 5{]} \\
$f_{16}(y) = 4 y_1^2 - 2.1 y_1^4 + \frac{1}{3} y_1^6 + y_1 y_2 - 4 y_2^2 + 4 y_2^4 $
    & 30, 50, 100                   &
    {[}-5, 5{]} \\
$f_{17}(y) = \left( y_2 - \frac{5.1}{4\pi^2} y_1^2 + \frac{5}{\pi} y_1 - 6 \right)^2 + 10 \left( 1 - \frac{1}{8\pi} \right) \cos y_1 + 10$
    & 30, 50, 100                   &
    {[}-5, 5{]} \\
$f_{18}(y) = [1 + (y_1 + y_2 + 1)^2] (19 - 14 y_1 + 3 y_1^2 - 14 y_2 + 6 y_1 y_2 + 3 y_2^2)$\\ \hspace{1cm}$\times [30 + (2 y_1 - 3 y_2)^2 \times (18 - 32 y_1 + 12 y_1^2 + 48 y_2 - 36 y_1 y_2 + 27 y_2^2)]$
     & 30, 50, 100                   &
    {[}-2, 2{]} \\
$f_{19}(y) = -\sum_{i=1}^{4} c_i \exp \left( -\sum_{j=1}^{3} a_{ij} (y_j - p_{ij})^2 \right)$
    & 30, 50, 100                   &
    {[}1, 3{]} \\
$f_{20}(y) = -\sum_{i=1}^{4} c_i \exp \left( -\sum_{j=1}^{6} a_{ij} (y_j - p_{ij})^2 \right)$
    & 30, 50, 100                   &
    {[}0, 1{]} \\
$f_{21}(y) = -\sum_{i=1}^{5} \left[ (y - a_i)(y - a_i)^T + c_i \right]^{-1}$
    & 30, 50, 100                   &
    {[}0, 10{]} \\
$f_{22}(y) = -\sum_{i=1}^{7} \left[ (y - a_i)(y - a_i)^T + c_i \right]^{-1}$
    & 30, 50, 100                   &
    {[}0, 10{]} \\
$f_{23}(y) = -\sum_{i=1}^{10} \left[ (y - a_i)(y - a_i)^T + c_i \right]^{-1}$
    & 30, 50, 100                   &
    {[}0, 10{]} \\
\hline             

\end{tabular}%
}
\end{table*}
\section{Experimental Results} 
The proposed algorithm is evaluated through simulation in MATLAB 2021a, running on Windows 11 with an AMD Ryzen 7 CPU and 16GB of RAM. It is benchmarked against 23 benchmark functions from the Congress on Evolutionary Computation (CEC 2017) suite \cite{Mallipeddi2010}. Details of these 23 benchmark functions are provided in Table \ref{table1}.

FA, SSO, HFASSO, and five hybrid algorithms are tested on each benchmark function ten times for 30, 50, and 100 dimensions. The average fitness value of HFASSON, based on ten executions, is compared with FA, SSO, and HFASSO  as well as with five selected hybrid algorithms. Each algorithm uses a population size of 100 with 1000 iterations. Additionally, the parameters used for each method are described in Table \ref{table2}.

The best fitness values and a comparison of HFASSON with WCMFO, HPSOGA, HSSOGSA, GWOCS, FAGA, SSO, FA, and HFASSO are summarized in Table \ref{table3}. The results show that HFASSON outperforms these algorithms for eight objective functions at 30D, and for nine objective functions at both 50D and 100D. A zero optimal value is attained for $f_{1}$ through $f_{4}$, as well as $f_{6}$, $f_{9}$, and $f_{11}$ across all three dimensions.

HFASSON achieves the optimal value more quickly than the other algorithms, even when their results are similar. The iteration counts for the best solution, presented in Table \ref{tablex}, reveal that HFASSON converges faster than the other methods. 
\begin{table}[]
\caption{Parameters of FA, SSO, HFASSO, HFASSON and CR-VANET\_HFASSON}
\label{table2}
\begin{tabular}{ll}
\hline
\multicolumn{1}{c}{\textbf{Name}}             & \multicolumn{1}{c}{\textbf{Value}}    \\ \hline
Simulator                                       & Matlab2021a                            \\ \hline
\multicolumn{2}{c}{\textbf{FA}}                                                        \\ \hline
Population size                                 & 30                                     \\
Numbers of   iterations/generations             & 1000                                   \\
Light absorption coefficient,   $\gamma$             & 1                                      \\
Attraction Coefficient Base   Value, $\beta_0$        & 2                                      \\
Mutation Coefficient, $\alpha$                         & 0.2                                    
      \\ \hline
\multicolumn{2}{c}{\textbf{SSO}}                                                       \\ \hline
Population size                                 & 30                                     \\
Numbers of   iterations/generations             & 1000                                   \\
Damping factor of velocity,   $\delta$                 & Rand (0, 1)                            \\
pH                                              & Rand (7, 14)                           \\
Temperature                                     & Rand (35.5, 38.5)                      \\ \hline
\multicolumn{2}{c}{\textbf{HFASSO/HFASSON}}                                            \\ \hline
Population size                                 & 30                                     \\
Light absorption coefficient,   $\gamma$               & 1                                      \\
Attraction Coefficient Base   Value, $\beta_0$          & 2                                      \\
Mutation Coefficient, $\alpha$                          & 0.2                                    \\

Numbers of   iterations/generations             & 1000                                   \\
Damping factor of velocity,   $\delta$                 & Rand (0, 1)                            \\
pH                                              & Rand (7, 14)                           \\
Temperature                                     & Rand (35.5, 38.5)                      \\ \hline
\multicolumn{2}{c}{\textbf{CR-VANET\_HFASSON}}                                         \\ \hline
Population size                                 & 30                                     \\
Light absorption coefficient,   $\gamma$              & 1                                      \\
Attraction Coefficient Base   Value, $\beta_0$          & 2                                      \\
Mutation Coefficient, $\alpha$                        & 0.2                                    \\

Numbers of   iterations/generations             & 1000                                   \\
Damping factor of velocity,   $\delta$                & Rand (0, 1)                            \\
pH                                              & Rand (7, 14)                           \\
Temperature                                     & Rand (35.5, 38.5)                      \\
Routing Algorithm                               & AODV                                   \\
Number of Vehicles                              & 20, 40, 60, 80, 100                    \\
Number of channels                              & 5                                      \\
Number of primary users                         & 5                                      \\
Number of secondary users                       & 6                                      \\
SNR in dB                                       & -10                                     \\
Maximum and minimum speed                      & 30 m/s and 10 m/s \\ \hline
\end{tabular}
\end{table}
\begin{table*}[htbp]
\centering
\caption{Numerical analysis of Best fitness value}
\label{table3}
\resizebox{\textwidth}{!}{%
\begin{tabular}{llllllllllllllllllllllllllll}
\hline
\multicolumn{1}{|l|}{\multirow{2}{*}{\textbf{Obj.Func}}} & \multicolumn{9}{c|}{\textbf{30D}}                                    & \multicolumn{9}{c|}{\textbf{50D}}                                & \multicolumn{9}{c|}{\textbf{100D}}                               \\ \cline{2-28} 
\multicolumn{1}{|l|}{}                                   & \multicolumn{1}{l|}{\textbf{WCMFO}} & \multicolumn{1}{l|}{\textbf{HPSOGA}}   & \multicolumn{1}{l|}{\textbf{HSSOGSA}} & \multicolumn{1}{l|}{\textbf{GWOCS}}     & \multicolumn{1}{l|}{\textbf{FAGA}}    & \multicolumn{1}{l|}{\textbf{FA}}      & \multicolumn{1}{l|}{\textbf{SSO}} & \multicolumn{1}{l|}{\textbf{HFASSO}}     & \multicolumn{1}{l|}{\textbf{HFASSON}}   & \multicolumn{1}{l|}{\textbf{WCMFO}} & \multicolumn{1}{l|}{\textbf{HPSOGA}}    & \multicolumn{1}{l|}{\textbf{HSSOGSA}} & \multicolumn{1}{l|}{\textbf{GWOCS}}    & \multicolumn{1}{l|}{\textbf{FAGA}}     & \multicolumn{1}{l|}{\textbf{FA}}         & \multicolumn{1}{l|}{\textbf{SSO}} & \multicolumn{1}{l|}{\textbf{HFASSO}}  & \multicolumn{1}{l|}{\textbf{HFASSON}}   & \multicolumn{1}{l|}{\textbf{WCMFO}} & \multicolumn{1}{l|}{\textbf{HPSOGA}}   & \multicolumn{1}{l|}{\textbf{HSSOGSA}} & \multicolumn{1}{l|}{\textbf{GWOCS}}  & \multicolumn{1}{l|}{\textbf{FAGA}}    & \multicolumn{1}{l|}{\textbf{FA}}       & \multicolumn{1}{l|}{\textbf{SSO}}      & \multicolumn{1}{l|}{\textbf{HFASSO}}   & \multicolumn{1}{l|}{\textbf{HFASSON}}   \\ \hline
\multicolumn{1}{|l|}{\textbf{1}}                         & \multicolumn{1}{l|}{19.01708}       & \multicolumn{1}{l|}{2.35E-37}          & \multicolumn{1}{l|}{0.031}            & \multicolumn{1}{l|}{8.52E-60}           & \multicolumn{1}{l|}{0.00019}          & \multicolumn{1}{l|}{1.8E-16}          & \multicolumn{1}{l|}{7.97E-258}    & \multicolumn{1}{l|}{\textbf{0}}          & \multicolumn{1}{l|}{\textbf{0}}         & \multicolumn{1}{l|}{18.1985}          & \multicolumn{1}{l|}{1.06E-20}           & \multicolumn{1}{l|}{112}              & \multicolumn{1}{l|}{1.53E-44}          & \multicolumn{1}{l|}{0.00175}           & \multicolumn{1}{l|}{7.007E-16}           & \multicolumn{1}{l|}{1.669E-189}   & \multicolumn{1}{l|}{\textbf{0}}       & \multicolumn{1}{l|}{\textbf{0}}         & \multicolumn{1}{l|}{20.7408}         & \multicolumn{1}{l|}{2.75E-08}          & \multicolumn{1}{l|}{8950}             & \multicolumn{1}{l|}{2.26E-30}        & \multicolumn{1}{l|}{0.0178}           & \multicolumn{1}{l|}{5.2E-15}           & \multicolumn{1}{l|}{1.2E-193}          & \multicolumn{1}{l|}{\textbf{0}}        & \multicolumn{1}{l|}{\textbf{0}}         \\ \hline
\multicolumn{1}{|l|}{\textbf{2}}                         & \multicolumn{1}{l|}{8.33489}        & \multicolumn{1}{l|}{7.22E-25}          & \multicolumn{1}{l|}{3.6E-09}          & \multicolumn{1}{l|}{0.04671}            & \multicolumn{1}{l|}{0.00134}          & \multicolumn{1}{l|}{5.9E-09}          & \multicolumn{1}{l|}{7.02E-147}    & \multicolumn{1}{l|}{\textbf{0}}          & \multicolumn{1}{l|}{\textbf{0}}         & \multicolumn{1}{l|}{8.1423}       & \multicolumn{1}{l|}{6.34E-16}           & \multicolumn{1}{l|}{19.5}             & \multicolumn{1}{l|}{6.50E-27}          & \multicolumn{1}{l|}{0.00569}           & \multicolumn{1}{l|}{1.37E-08}            & \multicolumn{1}{l|}{5.179E-117}   & \multicolumn{1}{l|}{\textbf{0}}       & \multicolumn{1}{l|}{\textbf{0}}         & \multicolumn{1}{l|}{7.9}       & 
\multicolumn{1}{l|}{9.34E-07}          & \multicolumn{1}{l|}{236}              & \multicolumn{1}{l|}{4.26E-19}        & \multicolumn{1}{l|}{0.04262}          & \multicolumn{1}{l|}{5.1E-08}           & \multicolumn{1}{l|}{8.8E-118}          & \multicolumn{1}{l|}{\textbf{0}}        & \multicolumn{1}{l|}{\textbf{0}}         \\ \hline
\multicolumn{1}{|l|}{\textbf{3}}                         & \multicolumn{1}{l|}{13.1885}        & \multicolumn{1}{l|}{0.12424}           & \multicolumn{1}{l|}{6130}             & \multicolumn{1}{l|}{4.64E-15}           & \multicolumn{1}{l|}{412.553}          & \multicolumn{1}{l|}{3.6E-09}          & \multicolumn{1}{l|}{1.28E-135}    & \multicolumn{1}{l|}{\textbf{0}}          & \multicolumn{1}{l|}{\textbf{0}}         & \multicolumn{1}{r|}{8.04E+01}       & \multicolumn{1}{r|}{1.81E+02}           & \multicolumn{1}{r|}{2.56E+04}         & \multicolumn{1}{r|}{1.49E-06}          & \multicolumn{1}{r|}{3.07E+03}          & \multicolumn{1}{l|}{0.0014126}           & \multicolumn{1}{l|}{6.7504E-66}   & \multicolumn{1}{l|}{\textbf{0}}       & \multicolumn{1}{l|}{\textbf{0}}         & \multicolumn{1}{l|}{18910.5125}         & \multicolumn{1}{l|}{6.82E+03}          & \multicolumn{1}{l|}{132000}           & \multicolumn{1}{l|}{6.74}            & \multicolumn{1}{l|}{25000}            & \multicolumn{1}{l|}{88.5115}           & \multicolumn{1}{l|}{2.58E-45}          & \multicolumn{1}{l|}{\textbf{0}}        & \multicolumn{1}{l|}{\textbf{0}}         \\ \hline
\multicolumn{1}{|l|}{\textbf{4}}                         & \multicolumn{1}{l|}{2.56121}        & \multicolumn{1}{l|}{0.22659}           & \multicolumn{1}{l|}{57.40876}         & \multicolumn{1}{l|}{0.0527}             & \multicolumn{1}{l|}{0.21732}          & \multicolumn{1}{l|}{3.48813}          & \multicolumn{1}{l|}{4E-100}       & \multicolumn{1}{l|}{\textbf{0}}          & \multicolumn{1}{l|}{\textbf{0}}         & \multicolumn{1}{l|}{2.5905}       & \multicolumn{1}{l|}{5.19731}            & \multicolumn{1}{l|}{77.03731}         & \multicolumn{1}{l|}{1.491E-10}         & \multicolumn{1}{l|}{0.7223}            & \multicolumn{1}{l|}{17.167256}           & \multicolumn{1}{l|}{1.3039E-58}   & \multicolumn{1}{l|}{\textbf{0}}       & \multicolumn{1}{l|}{\textbf{0}}         & \multicolumn{1}{l|}{12.3787}       & \multicolumn{1}{l|}{13.05651}          & \multicolumn{1}{l|}{95.55235}         & \multicolumn{1}{l|}{5.84E-06}        & \multicolumn{1}{l|}{3.69252}          & \multicolumn{1}{l|}{58.5539}           & \multicolumn{1}{l|}{1.05E-25}          & \multicolumn{1}{l|}{\textbf{0}}        & \multicolumn{1}{l|}{\textbf{0}}         \\ \hline
\multicolumn{1}{|l|}{\textbf{5}}                         & \multicolumn{1}{l|}{9013.4}         & \multicolumn{1}{l|}{29.21276}          & \multicolumn{1}{l|}{65.25057}         & \multicolumn{1}{l|}{\textbf{26.61761}}  & \multicolumn{1}{l|}{61.9829}          & \multicolumn{1}{l|}{32.6514}          & \multicolumn{1}{l|}{28.850349}    & \multicolumn{1}{l|}{28.9419206}          & \multicolumn{1}{l|}{28.674485}          & \multicolumn{1}{l|}{13153.0067}       & \multicolumn{1}{l|}{75.49686}           & \multicolumn{1}{l|}{140.84067}        & \multicolumn{1}{l|}{\textbf{47.34761}} & \multicolumn{1}{l|}{116.368}           & \multicolumn{1}{l|}{49.240324}           & \multicolumn{1}{l|}{48.2748122}   & \multicolumn{1}{l|}{48.9477}          & \multicolumn{1}{l|}{48.474372}          & \multicolumn{1}{l|}{10025.6033}       & \multicolumn{1}{l|}{268.5227}          & \multicolumn{1}{l|}{305000}           & \multicolumn{1}{l|}{\textbf{97.3}}   & \multicolumn{1}{l|}{1.6E+08}          & \multicolumn{1}{l|}{152.19}            & \multicolumn{1}{l|}{98.38423}          & \multicolumn{1}{l|}{98.93674}          & \multicolumn{1}{l|}{97.99341}           \\ \hline
\multicolumn{1}{|l|}{\textbf{6}}                         & \multicolumn{1}{l|}{18.40519}       & \multicolumn{1}{l|}{\textbf{0}}        & \multicolumn{1}{l|}{0.000603}         & \multicolumn{1}{l|}{0.55662}            & \multicolumn{1}{l|}{0.00049}          & \multicolumn{1}{l|}{1.9E-16}          & \multicolumn{1}{l|}{4.0798626}    & \multicolumn{1}{l|}{6.44386761}          & \multicolumn{1}{l|}{\textbf{0}}         & \multicolumn{1}{l|}{19.3321}          & \multicolumn{1}{l|}{8.56E-21}           & \multicolumn{1}{l|}{98.1}             & \multicolumn{1}{l|}{2.46}              & \multicolumn{1}{l|}{0.00092}           & \multicolumn{1}{l|}{6.29E-16}            & \multicolumn{1}{l|}{9.8171504}    & \multicolumn{1}{l|}{11.43502}         & \multicolumn{1}{l|}{\textbf{0}}         & \multicolumn{1}{l|}{24.2339}          & \multicolumn{1}{l|}{2.69E-08}          & \multicolumn{1}{l|}{5500}             & \multicolumn{1}{l|}{9.23}            & \multicolumn{1}{l|}{0.0153}           & \multicolumn{1}{l|}{5.2E-15}           & \multicolumn{1}{l|}{21.74713}          & \multicolumn{1}{l|}{23.49201}          & \multicolumn{1}{l|}{\textbf{0}}         \\ \hline
\multicolumn{1}{|l|}{\textbf{7}}                         & \multicolumn{1}{l|}{902.6863}       & \multicolumn{1}{l|}{0.00305}           & \multicolumn{1}{l|}{0.08336}          & \multicolumn{1}{l|}{0.0661}             & \multicolumn{1}{l|}{0.00483}          & \multicolumn{1}{l|}{0.00176}          & \multicolumn{1}{l|}{0.0002828}    & \multicolumn{1}{l|}{\textbf{1.9216E-05}} & \multicolumn{1}{l|}{63.98229}           & \multicolumn{1}{l|}{2667.5656}           & \multicolumn{1}{l|}{0.00739}            & \multicolumn{1}{l|}{0.27575}          & \multicolumn{1}{l|}{0.00129}           & \multicolumn{1}{l|}{0.0152}            & \multicolumn{1}{l|}{0.0027721}           & \multicolumn{1}{l|}{6.3413E-05}   & \multicolumn{1}{l|}{\textbf{2.6E-05}} & \multicolumn{1}{l|}{0.000158}           & \multicolumn{1}{l|}{10856.3195}           & \multicolumn{1}{l|}{0.0702}            & \multicolumn{1}{l|}{1.83}             & \multicolumn{1}{l|}{0.00205}         & \multicolumn{1}{l|}{0.0664}           & \multicolumn{1}{l|}{0.00887}           & \multicolumn{1}{l|}{\textbf{6.76E-05}} & \multicolumn{1}{l|}{1037.486}          & \multicolumn{1}{l|}{155.90751}          \\ \hline
\multicolumn{1}{|l|}{\textbf{8}}                         & \multicolumn{1}{l|}{-10272.2}       & \multicolumn{1}{l|}{-11200}            & \multicolumn{1}{l|}{-7480}            & \multicolumn{1}{l|}{\textbf{-11844.05}} & \multicolumn{1}{l|}{-10900}           & \multicolumn{1}{l|}{-8845.8}          & \multicolumn{1}{l|}{-5817.995}    & \multicolumn{1}{l|}{-2390.7943}          & \multicolumn{1}{l|}{-2137.0801}         & \multicolumn{1}{l|}{-15666.3835}         & \multicolumn{1}{l|}{-15600}             & \multicolumn{1}{l|}{-12300}           & \multicolumn{1}{l|}{\textbf{-20000}}   & \multicolumn{1}{l|}{-18100}            & \multicolumn{1}{l|}{-14469.391}          & \multicolumn{1}{l|}{-9287.3252}   & \multicolumn{1}{l|}{-3011.16}         & \multicolumn{1}{l|}{-3200.6974}         & \multicolumn{1}{l|}{-30345.9248}         & \multicolumn{1}{l|}{-25600}            & \multicolumn{1}{l|}{-23300}           & \multicolumn{1}{l|}{\textbf{-40800}} & \multicolumn{1}{l|}{-35800}           & \multicolumn{1}{l|}{-28139}            & \multicolumn{1}{l|}{-17843}            & \multicolumn{1}{l|}{-4806.37}          & \multicolumn{1}{l|}{-4256.792}          \\ \hline
\multicolumn{1}{|l|}{\textbf{9}}                         & \multicolumn{1}{l|}{47.02474}       & \multicolumn{1}{l|}{6.06925}           & \multicolumn{1}{l|}{105.46518}        & \multicolumn{1}{l|}{2.24549}            & \multicolumn{1}{l|}{7.3E-05}          & \multicolumn{1}{l|}{59.6974}          & \multicolumn{1}{l|}{\textbf{0}}   & \multicolumn{1}{l|}{\textbf{0}}          & \multicolumn{1}{l|}{\textbf{0}}         & \multicolumn{1}{l|}{103.2495}            & \multicolumn{1}{l|}{27.1}               & \multicolumn{1}{l|}{234}              & \multicolumn{1}{l|}{0.607}             & \multicolumn{1}{l|}{0.00036}           & \multicolumn{1}{l|}{136.90583}           & \multicolumn{1}{l|}{\textbf{0}}   & \multicolumn{1}{l|}{\textbf{0}}       & \multicolumn{1}{l|}{\textbf{0}}         & \multicolumn{1}{l|}{839.7641}            & \multicolumn{1}{l|}{87.6}              & \multicolumn{1}{l|}{472}              & \multicolumn{1}{l|}{0.474}           & \multicolumn{1}{l|}{0.0105}           & \multicolumn{1}{l|}{302.964}           & \multicolumn{1}{l|}{\textbf{0}}        & \multicolumn{1}{l|}{\textbf{0}}        & \multicolumn{1}{l|}{\textbf{0}}         \\ \hline
\multicolumn{1}{|l|}{\textbf{10}}                        & \multicolumn{1}{l|}{16.70351}       & \multicolumn{1}{l|}{0.0755}            & \multicolumn{1}{l|}{7.76}             & \multicolumn{1}{l|}{0.0147}             & \multicolumn{1}{l|}{0.001}            & \multicolumn{1}{l|}{3.3E-09}          & \multicolumn{1}{l|}{3.73E-15}     & \multicolumn{1}{l|}{\textbf{8.8818E-16}} & \multicolumn{1}{l|}{\textbf{8.882E-16}} & \multicolumn{1}{l|}{19.8444}           & \multicolumn{1}{l|}{6.928E-15}          & \multicolumn{1}{l|}{17.4}             & \multicolumn{1}{l|}{2.52E-14}          & \multicolumn{1}{l|}{0.0064}            & \multicolumn{1}{l|}{4.728E-09}           & \multicolumn{1}{l|}{8.8818E-16}   & \multicolumn{1}{l|}{8.88E-16}         & \multicolumn{1}{l|}{\textbf{8.438E-16}} & \multicolumn{1}{r|}{2.00E+01}       & \multicolumn{1}{r|}{9.54E-01}          & \multicolumn{1}{r|}{1.97E+01}         & \multicolumn{1}{r|}{1.75E-14}        & \multicolumn{1}{r|}{1.53E-02}         & \multicolumn{1}{l|}{9.2E-09}           & \multicolumn{1}{l|}{\textbf{8.88E-16}} & \multicolumn{1}{l|}{\textbf{8.88E-16}} & \multicolumn{1}{l|}{\textbf{8.882E-16}} \\ \hline
\multicolumn{1}{|l|}{\textbf{11}}                        & \multicolumn{1}{l|}{0.4828}        & \multicolumn{1}{l|}{0.00148}           & \multicolumn{1}{l|}{1.08099}          & \multicolumn{1}{l|}{0.00265}            & \multicolumn{1}{l|}{0.02083}          & \multicolumn{1}{l|}{3.1E-10}          & \multicolumn{1}{l|}{\textbf{0}}   & \multicolumn{1}{l|}{\textbf{0}}          & \multicolumn{1}{l|}{\textbf{0}}         & \multicolumn{1}{l|}{229.9926}       & \multicolumn{1}{l|}{0.00148}            & \multicolumn{1}{l|}{19.86612}         & \multicolumn{1}{l|}{\textbf{0}}        & \multicolumn{1}{l|}{0.00666}           & \multicolumn{1}{l|}{0.0029552}           & \multicolumn{1}{l|}{\textbf{0}}   & \multicolumn{1}{l|}{\textbf{0}}       & \multicolumn{1}{l|}{\textbf{0}}         & \multicolumn{1}{l|}{0.2764}            & \multicolumn{1}{l|}{0.00246}           & \multicolumn{1}{l|}{257}              & \multicolumn{1}{l|}{\textbf{0}}      & \multicolumn{1}{l|}{0.0163}           & \multicolumn{1}{l|}{0.00074}           & \multicolumn{1}{l|}{\textbf{0}}        & \multicolumn{1}{l|}{\textbf{0}}        & \multicolumn{1}{l|}{\textbf{0}}         \\ \hline
\multicolumn{1}{|l|}{\textbf{12}}                        & \multicolumn{1}{l|}{0.55837}        & \multicolumn{1}{l|}{\textbf{1.57E-32}} & \multicolumn{1}{l|}{9.52337}          & \multicolumn{1}{l|}{0.03873}            & \multicolumn{1}{l|}{5.4E-06}          & \multicolumn{1}{l|}{0.0933}           & \multicolumn{1}{l|}{0.400079}     & \multicolumn{1}{l|}{1.2149839}           & \multicolumn{1}{l|}{3.576E-32}          & \multicolumn{1}{l|}{0.2869}       & \multicolumn{1}{l|}{3.47E-22}           & \multicolumn{1}{l|}{18.5}             & \multicolumn{1}{l|}{0.0964}            & \multicolumn{1}{l|}{1.6E-06}           & \multicolumn{1}{l|}{0.1431419}           & \multicolumn{1}{l|}{0.74908074}   & \multicolumn{1}{l|}{1.200264}         & \multicolumn{1}{l|}{\textbf{1.173E-32}} & \multicolumn{1}{l|}{0.5496}       & \multicolumn{1}{l|}{0.04667}           & \multicolumn{1}{l|}{390000}           & \multicolumn{1}{l|}{0.251}           & \multicolumn{1}{l|}{1.3E-05}          & \multicolumn{1}{l|}{0.19342}           & \multicolumn{1}{l|}{0.901571}          & \multicolumn{1}{l|}{1.162765}          & \multicolumn{1}{l|}{\textbf{7.052E-33}} \\ \hline
\multicolumn{1}{|l|}{\textbf{13}}                        & \multicolumn{1}{l|}{1.88644}        & \multicolumn{1}{l|}{\textbf{1.35E-32}} & \multicolumn{1}{l|}{22.67375}         & \multicolumn{1}{l|}{0.67616}            & \multicolumn{1}{l|}{1.3E-05}          & \multicolumn{1}{l|}{8.4E-18}          & \multicolumn{1}{l|}{2.2542496}    & \multicolumn{1}{l|}{2.98826819}          & \multicolumn{1}{l|}{2.9316923}          & \multicolumn{1}{l|}{1.9065}       & \multicolumn{1}{l|}{0.0011}             & \multicolumn{1}{l|}{194}              & \multicolumn{1}{l|}{1.7}               & \multicolumn{1}{l|}{0.00013}           & \multicolumn{1}{l|}{\textbf{2.452E-17}}  & \multicolumn{1}{l|}{4.6308651}    & \multicolumn{1}{l|}{4.979216}         & \multicolumn{1}{l|}{4.9217444}          & \multicolumn{1}{l|}{4.6882}       & \multicolumn{1}{l|}{0.067775}          & \multicolumn{1}{l|}{3370000}          & \multicolumn{1}{l|}{6.06}            & \multicolumn{1}{l|}{\textbf{0.00072}} & \multicolumn{1}{l|}{0.0033}            & \multicolumn{1}{l|}{9.550715}          & \multicolumn{1}{l|}{9.992374}          & \multicolumn{1}{l|}{9.8835503}          \\ \hline
\multicolumn{1}{|l|}{\textbf{14}}                        & \multicolumn{1}{l|}{2.9821}         & \multicolumn{1}{l|}{\textbf{0.998}}    & \multicolumn{1}{l|}{5.46951}          & \multicolumn{1}{l|}{5.98017}            & \multicolumn{1}{l|}{\textbf{0.998}}   & \multicolumn{1}{l|}{0.998}            & \multicolumn{1}{l|}{8.2075928}    & \multicolumn{1}{l|}{11.8054744}          & \multicolumn{1}{l|}{12.512941}          & \multicolumn{1}{l|}{2.9821}            & \multicolumn{1}{l|}{0.998}              & \multicolumn{1}{l|}{5.6865}           & \multicolumn{1}{l|}{\textbf{-1990}}    & \multicolumn{1}{l|}{0.998}             & \multicolumn{1}{l|}{0.9980038}           & \multicolumn{1}{l|}{3.3635095}    & \multicolumn{1}{l|}{9.940117}         & \multicolumn{1}{l|}{12.140736}          & \multicolumn{1}{l|}{4.9322}            & \multicolumn{1}{l|}{0.998}             & \multicolumn{1}{l|}{3.66}             & \multicolumn{1}{l|}{466}             & \multicolumn{1}{l|}{\textbf{0.998}}   & \multicolumn{1}{l|}{0.998}             & \multicolumn{1}{l|}{1.990103}          & \multicolumn{1}{l|}{12.48576}          & \multicolumn{1}{l|}{12.010137}          \\ \hline
\multicolumn{1}{|l|}{\textbf{15}}                        & \multicolumn{1}{l|}{0.01729}        & \multicolumn{1}{l|}{0.0925}            & \multicolumn{1}{l|}{0.00288}          & \multicolumn{1}{l|}{0.0926}             & \multicolumn{1}{l|}{0.00076}          & \multicolumn{1}{l|}{\textbf{0.00031}} & \multicolumn{1}{l|}{0.0003397}    & \multicolumn{1}{l|}{0.02287112}          & \multicolumn{1}{l|}{0.000815}           & \multicolumn{1}{l|}{0.017096}           & \multicolumn{1}{l|}{0.00231}            & \multicolumn{1}{l|}{0.0028}           & \multicolumn{1}{l|}{\textbf{0.00039}}  & \multicolumn{1}{l|}{0.00509}           & \multicolumn{1}{l|}{0.0003991}           & \multicolumn{1}{l|}{0.00043513}   & \multicolumn{1}{l|}{0.011596}         & \multicolumn{1}{l|}{0.0006999}          & \multicolumn{1}{l|}{0.0605}           & \multicolumn{1}{l|}{\textbf{0.0003}}   & \multicolumn{1}{l|}{0.00664}          & \multicolumn{1}{l|}{0.000307}        & \multicolumn{1}{l|}{0.0055}           & \multicolumn{1}{l|}{0.00031}           & \multicolumn{1}{l|}{0.00034}           & \multicolumn{1}{l|}{0.009812}          & \multicolumn{1}{l|}{0.0007146}          \\ \hline
\multicolumn{1}{|l|}{\textbf{16}}                        & \multicolumn{1}{l|}{28.77476}       & \multicolumn{1}{l|}{-1.0316}           & \multicolumn{1}{l|}{-1.0316}          & \multicolumn{1}{l|}{-1.0316}            & \multicolumn{1}{l|}{-1.0316}          & \multicolumn{1}{l|}{\textbf{-1.0316}} & \multicolumn{1}{l|}{0.000406}     & \multicolumn{1}{l|}{-1}                  & \multicolumn{1}{l|}{-1.0076266}         & \multicolumn{1}{l|}{-0.2155}       & \multicolumn{1}{l|}{-1.0316}            & \multicolumn{1}{l|}{-1.0316}          & \multicolumn{1}{l|}{-1.0316}           & \multicolumn{1}{l|}{-1.0316}           & \multicolumn{1}{l|}{\textbf{-1.0316285}} & \multicolumn{1}{l|}{-1.0315161}   & \multicolumn{1}{l|}{-1.00075}         & \multicolumn{1}{l|}{-0.9999287}         & \multicolumn{1}{l|}{-0.2155}        & \multicolumn{1}{l|}{-1.0316}           & \multicolumn{1}{l|}{-1.0316}          & \multicolumn{1}{l|}{-1.0316}         & \multicolumn{1}{l|}{-1.03156}         & \multicolumn{1}{l|}{\textbf{-1.03163}} & \multicolumn{1}{l|}{-1.03159}          & \multicolumn{1}{l|}{-1.00004}          & \multicolumn{1}{l|}{-1.001464}          \\ \hline
\multicolumn{1}{|l|}{\textbf{17}}                        & \multicolumn{1}{l|}{16.3289}         & \multicolumn{1}{l|}{\textbf{0.3979}}   & \multicolumn{1}{l|}{\textbf{0.3979}}  & \multicolumn{1}{l|}{\textbf{0.3979}}    & \multicolumn{1}{l|}{\textbf{0.3979}}  & \multicolumn{1}{l|}{17.1781}          & \multicolumn{1}{l|}{0.3987835}    & \multicolumn{1}{l|}{1.52736576}          & \multicolumn{1}{l|}{0.9418662}          & \multicolumn{1}{l|}{16.32891}       & \multicolumn{1}{l|}{\textbf{0.3979}}    & \multicolumn{1}{l|}{\textbf{0.3979}}  & \multicolumn{1}{l|}{49.40371}          & \multicolumn{1}{l|}{\textbf{0.3979}}   & \multicolumn{1}{l|}{422.78872}           & \multicolumn{1}{l|}{675.829673}   & \multicolumn{1}{l|}{1067.574}         & \multicolumn{1}{l|}{153.6188}           & \multicolumn{1}{l|}{0.3979}           & \multicolumn{1}{l|}{\textbf{0.3979}}   & \multicolumn{1}{l|}{\textbf{0.3979}}  & \multicolumn{1}{l|}{51.34023}        & \multicolumn{1}{l|}{0.3979}           & \multicolumn{1}{l|}{858.905}           & \multicolumn{1}{l|}{1749.776}          & \multicolumn{1}{l|}{1396.371}          & \multicolumn{1}{l|}{232.84955}          \\ \hline
\multicolumn{1}{|l|}{\textbf{18}}                        & \multicolumn{1}{l|}{44427}          & \multicolumn{1}{l|}{\textbf{3}}        & \multicolumn{1}{l|}{\textbf{3}}       & \multicolumn{1}{l|}{\textbf{3}}         & \multicolumn{1}{l|}{5.7}              & \multicolumn{1}{l|}{3}                & \multicolumn{1}{l|}{3.0000035}    & \multicolumn{1}{l|}{32.6247153}          & \multicolumn{1}{l|}{13.734564}          & \multicolumn{1}{l|}{33246.1856}          & \multicolumn{1}{l|}{\textbf{3}}         & \multicolumn{1}{l|}{\textbf{3}}       & \multicolumn{1}{l|}{\textbf{3}}        & \multicolumn{1}{l|}{5.70014}           & \multicolumn{1}{l|}{\textbf{3}}          & \multicolumn{1}{l|}{3.00000059}   & \multicolumn{1}{l|}{8.723104}         & \multicolumn{1}{l|}{8.3710409}          & \multicolumn{1}{l|}{44427.2626}          & \multicolumn{1}{l|}{\textbf{3}}        & \multicolumn{1}{l|}{\textbf{3}}       & \multicolumn{1}{l|}{\textbf{3}}      & \multicolumn{1}{l|}{8.4}              & \multicolumn{1}{l|}{\textbf{3}}        & \multicolumn{1}{l|}{\textbf{3}}        & \multicolumn{1}{l|}{4.270978}          & \multicolumn{1}{l|}{9.6462625}          \\ \hline
\multicolumn{1}{|l|}{\textbf{19}}                        & \multicolumn{1}{l|}{-0.3005}        & \multicolumn{1}{l|}{\textbf{-3.8628}}  & \multicolumn{1}{l|}{\textbf{-3.8628}} & \multicolumn{1}{l|}{-3.86193}           & \multicolumn{1}{l|}{\textbf{-3.8628}} & \multicolumn{1}{l|}{-3.8628}          & \multicolumn{1}{l|}{-3.71623}     & \multicolumn{1}{l|}{-3.4771155}          & \multicolumn{1}{l|}{-3.5288511}         & \multicolumn{1}{l|}{-9.25837}       & \multicolumn{1}{l|}{\textbf{-58.55823}} & \multicolumn{1}{l|}{-58.49617}        & \multicolumn{1}{l|}{-3.86275}          & \multicolumn{1}{l|}{-56.0368}          & \multicolumn{1}{l|}{-3.8627821}          & \multicolumn{1}{l|}{-3.7951707}   & \multicolumn{1}{l|}{-3.50954}         & \multicolumn{1}{l|}{-0.7445502}         & \multicolumn{1}{l|}{-21.7698}       & \multicolumn{1}{l|}{\textbf{-122.324}} & \multicolumn{1}{l|}{-114.7087}        & \multicolumn{1}{l|}{-3.8613}         & \multicolumn{1}{l|}{-113.182}         & \multicolumn{1}{l|}{-3.8628}           & \multicolumn{1}{l|}{-3.81993}          & \multicolumn{1}{l|}{-3.69358}          & \multicolumn{1}{l|}{-3.56282}           \\ \hline
\multicolumn{1}{|l|}{\textbf{20}}                        & \multicolumn{1}{l|}{-0.03802}       & \multicolumn{1}{r|}{3.137602}          & \multicolumn{1}{l|}{-3.21499}         & \multicolumn{1}{l|}{-3.25567}           & \multicolumn{1}{l|}{\textbf{-3.2982}} & \multicolumn{1}{l|}{-3.2863}          & \multicolumn{1}{l|}{-2.610521}    & \multicolumn{1}{l|}{-2.045542}           & \multicolumn{1}{l|}{-1.7258886}         & \multicolumn{1}{l|}{-0.04023}       & \multicolumn{1}{l|}{-3.09129}           & \multicolumn{1}{l|}{-3.25066}         & \multicolumn{1}{l|}{-3.26462}          & \multicolumn{1}{l|}{\textbf{-3.29822}} & \multicolumn{1}{l|}{\textbf{-3.2982165}} & \multicolumn{1}{l|}{-2.6059053}   & \multicolumn{1}{l|}{-1.76084}         & \multicolumn{1}{l|}{-2.538E-30}         & \multicolumn{1}{l|}{-0.05012}       & \multicolumn{1}{l|}{-3.145819}         & \multicolumn{1}{l|}{-3.27444}         & \multicolumn{1}{l|}{-3.30334}        & \multicolumn{1}{l|}{-3.25066}         & \multicolumn{1}{l|}{\textbf{-3.322}}   & \multicolumn{1}{l|}{-2.79217}          & \multicolumn{1}{l|}{-2.186}            & \multicolumn{1}{l|}{-2.075261}          \\ \hline
\multicolumn{1}{|l|}{\textbf{21}}                        & \multicolumn{1}{l|}{-0.2928}        & \multicolumn{1}{r|}{-2.67762}          & \multicolumn{1}{l|}{-4.41351}         & \multicolumn{1}{l|}{-8.63704}           & \multicolumn{1}{l|}{-8.6591}          & \multicolumn{1}{l|}{\textbf{-10.153}} & \multicolumn{1}{l|}{-5.232448}    & \multicolumn{1}{l|}{-1.2530012}          & \multicolumn{1}{l|}{-0.5160428}         & \multicolumn{1}{l|}{-0.26657}       & \multicolumn{1}{l|}{-1.59926}           & \multicolumn{1}{l|}{-4.89709}         & \multicolumn{1}{l|}{-7.11214}          & \multicolumn{1}{l|}{-5.66054}          & \multicolumn{1}{l|}{\textbf{-10.1532}}   & \multicolumn{1}{l|}{-3.4857049}   & \multicolumn{1}{l|}{-1.18444}         & \multicolumn{1}{l|}{-0.2146841}         & \multicolumn{1}{l|}{-0.26816}       & \multicolumn{1}{l|}{-2.39889}          & \multicolumn{1}{l|}{-5.90757}         & \multicolumn{1}{l|}{-8.12274}        & \multicolumn{1}{l|}{-5.90227}         & \multicolumn{1}{l|}{\textbf{-9.3087}}  & \multicolumn{1}{l|}{-3.13116}          & \multicolumn{1}{l|}{-1.23439}          & \multicolumn{1}{l|}{-0.812364}          \\ \hline
\multicolumn{1}{|l|}{\textbf{22}}                        & \multicolumn{1}{l|}{-0.3373}        & \multicolumn{1}{l|}{-6.81511}          & \multicolumn{1}{l|}{-6.58601}         & \multicolumn{1}{l|}{-8.81621}           & \multicolumn{1}{l|}{-4.7711}          & \multicolumn{1}{l|}{\textbf{-10.153}} & \multicolumn{1}{l|}{-4.451639}    & \multicolumn{1}{l|}{-1.1623998}          & \multicolumn{1}{l|}{-0.5972496}         & \multicolumn{1}{l|}{-0.27241}       & \multicolumn{1}{l|}{-2.13584}           & \multicolumn{1}{l|}{-5.1553}          & \multicolumn{1}{l|}{-8.3901}           & \multicolumn{1}{l|}{-2.17344}          & \multicolumn{1}{l|}{\textbf{-10.1532}}   & \multicolumn{1}{l|}{-2.9272848}   & \multicolumn{1}{l|}{-1.33837}         & \multicolumn{1}{l|}{-0.206796}          & \multicolumn{1}{l|}{-0.27206}       & \multicolumn{1}{l|}{-2.66718}          & \multicolumn{1}{l|}{-3.88867}         & \multicolumn{1}{l|}{-8.6325}         & \multicolumn{1}{l|}{-8.1076}          & \multicolumn{1}{l|}{\textbf{-10.1532}} & \multicolumn{1}{l|}{-4.05509}          & \multicolumn{1}{l|}{-2.30444}          & \multicolumn{1}{l|}{-1.04955}           \\ \hline
\multicolumn{1}{|l|}{\textbf{23}}                        & \multicolumn{1}{l|}{-0.40621}       & \multicolumn{1}{l|}{-5.31185}          & \multicolumn{1}{l|}{-6.32545}         & \multicolumn{1}{l|}{\textbf{-9.9998}}   & \multicolumn{1}{l|}{-9.0998}          & \multicolumn{1}{l|}{-4.4973}          & \multicolumn{1}{l|}{-4.26346}     & \multicolumn{1}{l|}{-1.1829832}          & \multicolumn{1}{l|}{-1.464341}          & \multicolumn{1}{l|}{-0.356}         & \multicolumn{1}{l|}{0}                  & \multicolumn{1}{l|}{-1.66}            & \multicolumn{1}{l|}{\textbf{-11.8}}    & \multicolumn{1}{l|}{-4.92}             & \multicolumn{1}{l|}{-2.5497008}          & \multicolumn{1}{l|}{-0.778286}    & \multicolumn{1}{l|}{-1.45403}         & \multicolumn{1}{l|}{-1.2191054}         & \multicolumn{1}{l|}{-0.198}         & \multicolumn{1}{l|}{0}                 & \multicolumn{1}{l|}{-0.645}           & \multicolumn{1}{l|}{\textbf{-10.6}}  & \multicolumn{1}{l|}{-0.912}           & \multicolumn{1}{l|}{-1.2158}           & \multicolumn{1}{l|}{-0.37527}          & \multicolumn{1}{l|}{-0.41218}          & \multicolumn{1}{l|}{-0.098451}          \\ \hline
\multicolumn{28}{l}{\textbf{*Bold font represents the best value.}}                                                          \end{tabular}%
}
\end{table*}

\begin{table}[ht]
\caption{Comparison of the iteration value at optimal solution.}
\label{tablex}
\resizebox{\columnwidth}{!}{%
\begin{tabular}{|p{2.5cm}|p{2.5cm}|p{1.5cm}|p{1.5cm}|p{1.5cm}|}
\hline
 &  & \multicolumn{3}{c|}{\shortstack{\textbf{Iteration value} \\ \textbf{at the optimal solution}}} \\ \cline{3-5}  
\multirow{-2}{*}{\textbf{Objective function}} & \multirow{-2}{*}{\textbf{Algorithm}} & \multicolumn{1}{c|}{\textbf{30D}} & \multicolumn{1}{c|}{\textbf{50D}} & \textbf{100D} \\ \hline
 & HFASSO & \multicolumn{1}{c|}{258} & \multicolumn{1}{c|}{264} & 270 \\ \cline{2-5} 
\multirow{-2}{*}{\textbf{f1}} & HFASSON & \multicolumn{1}{c|}{\textbf{142}} & \multicolumn{1}{c|}{\textbf{145}} & \textbf{148} \\ \hline
 & HFASSO & \multicolumn{1}{c|}{477} & \multicolumn{1}{c|}{486} & 498 \\ \cline{2-5} 
\multirow{-2}{*}{\textbf{f2}} & HFASSON & \multicolumn{1}{c|}{\textbf{268}} & \multicolumn{1}{c|}{\textbf{275}} & \textbf{281} \\ \hline
 & HFASSO & \multicolumn{1}{c|}{268} & \multicolumn{1}{c|}{272} & 279 \\ \cline{2-5} 
\multirow{-2}{*}{\textbf{f3}} & HFASSON & \multicolumn{1}{c|}{\textbf{147}} & \multicolumn{1}{c|}{\textbf{151}} & \textbf{154} \\ \hline
 & HFASSO & \multicolumn{1}{c|}{492} & \multicolumn{1}{c|}{503} & 513 \\ \cline{2-5} 
\multirow{-2}{*}{\textbf{f4}} & HFASSON & \multicolumn{1}{c|}{\textbf{279}} & \multicolumn{1}{c|}{\textbf{284}} & \textbf{284} \\ \hline
 & HPSOGA & \multicolumn{1}{c|}{861} & \multicolumn{1}{c|}{\cellcolor[HTML]{A6A6A6}} & \cellcolor[HTML]{A6A6A6} \\ \cline{2-5} 
\multirow{-2}{*}{\textbf{f6}} & HFASSON & \multicolumn{1}{c|}{\textbf{20}} & \multicolumn{1}{c|}{\textbf{21}} & \textbf{22} \\ \hline
 & SSO & \multicolumn{1}{c|}{95} & \multicolumn{1}{c|}{113} & 119 \\ \cline{2-5} 
 & HFASSO & \multicolumn{1}{c|}{18} & \multicolumn{1}{c|}{18} & 19 \\ \cline{2-5} 
\multirow{-3}{*}{\textbf{f9}} & HFASSON & \multicolumn{1}{c|}{\textbf{11}} & \multicolumn{1}{c|}{\textbf{12}} & \textbf{12} \\ \hline
 & SSO & \multicolumn{1}{c|}{\cellcolor[HTML]{A6A6A6}} & \multicolumn{1}{c|}{566} & 837 \\ \cline{2-5} 
 & HFASSO & \multicolumn{1}{c|}{32} & \multicolumn{1}{c|}{32} & 33 \\ \cline{2-5} 
\multirow{-3}{*}{\textbf{f10}} & HFASSON & \multicolumn{1}{c|}{\textbf{18}} & \multicolumn{1}{c|}{\textbf{19}} & \textbf{20} \\ \hline
 & SSO & \multicolumn{1}{c|}{115} & \multicolumn{1}{c|}{131} & 153 \\ \cline{2-5} 
 & HFASSO & \multicolumn{1}{c|}{20} & \multicolumn{1}{c|}{21} & 21 \\ \cline{2-5} 
\multirow{-3}{*}{\textbf{f11}} & HFASSON & \multicolumn{1}{c|}{\textbf{13}} & \multicolumn{1}{c|}{\textbf{13}} & \textbf{14} \\ \hline
\multicolumn{5}{l}{\textbf{*Bold values represent lowest iteration value at which optimal solution obtained.}}
\end{tabular}%
}
\end{table}
\section{Statistical Analysis}
A statistical test is performed to examine the variations in the distribution of results. The statistical analysis of the proposed HFASSON is conducted by comparing it with competing algorithms using the Friedman test. In this ranking, 
 HFASSON ranks highest for most of the objective functions. The summary in Table \ref{table8} shows that HFASSON achieves the first rank for 9 objective functions each in the 30D, 50D, and 100D settings.

\begin{table*}
\centering
\caption{Ranking of FA, SSO, HFASSO, HFASSON, and the selected state-of-the-art algorithms using Friedman test}
\label{table8}
\resizebox{\textwidth}{!}{%
\begin{tabular}{cccccccccccccccccccccccc} \\ \hline
\textbf{Rank   Obtained} & \textbf{1} & \textbf{2} & \textbf{3} & \textbf{4} & \textbf{5} & \textbf{6} & \textbf{7} & \textbf{8} & \textbf{9} & \textbf{10} & \textbf{11} & \textbf{12} & \textbf{13} & \textbf{14} & \textbf{15} & \textbf{16} & \textbf{17} & \textbf{18} & \textbf{19} & \textbf{20} & \textbf{21} & \textbf{22} & \textbf{23} \\ \hline
\multicolumn{24}{c}{\textbf{30D}} \\ \hline
\textbf{1} & \cellcolor[HTML]{92D050}HFASSON & \cellcolor[HTML]{92D050}HFASSON & \cellcolor[HTML]{92D050}HFASSON & \cellcolor[HTML]{92D050}HFASSON & GWOCS & \cellcolor[HTML]{92D050}HFASSON & HFASSO & HPSOGA & \cellcolor[HTML]{92D050}HFASSON & \cellcolor[HTML]{92D050}HFASSON & \cellcolor[HTML]{92D050}HFASSON & \cellcolor[HTML]{92D050}HFASSON & HPSOGA & FAGA & FA & FAGA & FAGA & FA & FAGA & FAGA & FA & FA & FAGA \\
\textbf{2} & HFASSO & HFASSO & HFASSO & HFASSO & HFASSON & HPSOGA & SSO & FAGA & HFASSO & HFASSO & HFASSO & HPSOGA & FA & HPSOGA & SSO & HPSOGA & HPSOGA & HPSOGA & HPSOGA & FA & FAGA & GWOCS & GWOCS \\
\textbf{3} & SSO & SSO & SSO & SSO & SSO & FA & GWOCS & WCMFO & SSO & SSO & SSO & FAGA & FAGA & FA & FAGA & FA & HSSOGSA & SSO & HSSOGSA & GWOCS & GWOCS & HPSOGA & HSSOGSA \\
\textbf{4} & GWOCS & HPSOGA & GWOCS & GWOCS & HFASSO & FAGA & FA & FA & FAGA & FA & FA & GWOCS & GWOCS & HSSOGSA & HFASSON & GWOCS & GWOCS & HSSOGSA & FA & HSSOGSA & HSSOGSA & HSSOGSA & HPSOGA \\
\textbf{5} & HPSOGA & FA & FA & FAGA & HPSOGA & GWOCS & HPSOGA & GWOCS & GWOCS & FAGA & GWOCS & FA & SSO & WCMFO & HSSOGSA & HSSOGSA & SSO & GWOCS & GWOCS & HPSOGA & SSO & FAGA & FA \\
\textbf{6} & FA & GWOCS & GWOCS & HPSOGA & FA & HSSOGSA & FAGA & HSSOGSA & HPSOGA & GWOCS & HPSOGA & SSO & WCMFO & GWOCS & HFASSO & HFASSON & HFASSON & FAGA & SSO & SSO & HPSOGA & SSO & SSO \\
\textbf{7} & FAGA & FAGA & WCMFO & WCMFO & FAGA & SSO & HSSOGSA & SSO & FA & HPSOGA & FAGA & HFASSO & HFASSON & SSO & WCMFO & HFASSO & HFASSO & HFASSON & HFASSON & HFASSO & HFASSO & HFASSO & HFASSON \\
\textbf{8} & HSSOGSA & HSSOGSA & FAGA & FA & HSSOGSA & HFASSO & HFASSON & HFASSO & WCMFO & HSSOGSA & WCMFO & WCMFO & HFASSO & HFASSO & HPSOGA & SSO & WCMFO & HFASSO & HFASSO & HFASSON & HFASSON & HFASSON & HFASSO \\
\textbf{9} & WCMFO & WCMFO & HSSOGSA & HSSOGSA & WCMFO & WCMFO & WCMFO & HFASSON & HSSOGSA & WCMFO & HSSOGSA & HSSOGSA & HSSOGSA & HFASSON & GWOCS & WCMFO & FA & WCMFO & WCMFO & WCMFO & WCMFO & WCMFO & WCMFO \\ \hline
\multicolumn{24}{c}{\textbf{50D}} \\ \hline
\textbf{1} & \cellcolor[HTML]{92D050}HFASSON & \cellcolor[HTML]{92D050}HFASSON & \cellcolor[HTML]{92D050}HFASSON & \cellcolor[HTML]{92D050}HFASSON & GWOCS & \cellcolor[HTML]{92D050}HFASSON & SSO & FAGA & \cellcolor[HTML]{92D050}HFASSON & \cellcolor[HTML]{92D050}HFASSON & \cellcolor[HTML]{92D050}HFASSON & \cellcolor[HTML]{92D050}HFASSON & FA & GWOCS & FA & FA & HPSOGA & HPSOGA & HPSOGA & FAGA & FA & FA & GWOCS \\
\textbf{2} & HFASSO & HFASSO & HFASSO & HFASSO & SSO & HPSOGA & HFASSON & HPSOGA & HFASSO & HFASSO & HFASSO & HPSOGA & FAGA & FAGA & SSO & HPSOGA & HSSOGSA & FA & HSSOGSA & FA & FAGA & GWOCS & FAGA \\
\textbf{3} & SSO & GWOCS & SSO & SSO & HFASSON & FA & HFASSO & FA & SSO & SSO & SSO & FAGA & HPSOGA & HPSOGA & GWOCS & FAGA & FAGA & HSSOGSA & FAGA & GWOCS & GWOCS & HSSOGSA & FA \\
\textbf{4} & GWOCS & SSO & GWOCS & GWOCS & HFASSO & FAGA & GWOCS & GWOCS & FAGA & HPSOGA & GWOCS & GWOCS & GWOCS & FA & HFASSON & GWOCS & WCMFO & GWOCS & WCMFO & HSSOGSA & HSSOGSA & SSO & HSSOGSA \\
\textbf{5} & HPSOGA & HPSOGA & FA & FAGA & FA & GWOCS & FA & WCMFO & GWOCS & GWOCS & HPSOGA & FA & SSO & SSO & HPSOGA & HSSOGSA & GWOCS & SSO & FA & HPSOGA & SSO & HPSOGA & HFASSO \\
\textbf{6} & FA & FA & HPSOGA & HPSOGA & HPSOGA & SSO & HPSOGA & HSSOGSA & HPSOGA & FA & FA & SSO & HFASSON & HSSOGSA & HSSOGSA & SSO & HFASSON & FAGA & GWOCS & SSO & HPSOGA & FAGA & HFASSON \\
\textbf{7} & FAGA & FAGA & FAGA & FA & FAGA & HFASSO & FAGA & SSO & FA & FAGA & FAGA & HFASSO & HFASSO & HFASSO & FAGA & HFASSO & FA & HFASSON & SSO & HFASSO & HFASSO & HFASSO & SSO \\
\textbf{8} & HSSOGSA & HSSOGSA & HSSOGSA & WCMFO & HSSOGSA & HSSOGSA & HSSOGSA & HFASSON & HSSOGSA & HSSOGSA & HSSOGSA & HSSOGSA & HSSOGSA & HFASSON & HFASSO & HFASSON & SSO & HFASSO & HFASSO & WCMFO & WCMFO & WCMFO & WCMFO \\
\textbf{9} & WCMFO & WCMFO & WCMFO & HSSOGSA & WCMFO & WCMFO & WCMFO & HFASSO & WCMFO & WCMFO & WCMFO & WCMFO & WCMFO & WCMFO & WCMFO & WCMFO & HFASSO & WCMFO & HFASSON & HFASSON & HFASSON & HFASSON & HPSOGA \\ \hline
\multicolumn{24}{c}{\textbf{100D}} \\ \hline
\textbf{1} & \cellcolor[HTML]{92D050}HFASSON & \cellcolor[HTML]{92D050}HFASSON & \cellcolor[HTML]{92D050}HFASSON & \cellcolor[HTML]{92D050}HFASSON & GWOCS & \cellcolor[HTML]{92D050}HFASSON & SSO & FAGA & \cellcolor[HTML]{92D050}HFASSON & HFASSO & \cellcolor[HTML]{92D050}HFASSON & \cellcolor[HTML]{92D050}HFASSON & FAGA & HPSOGA & HPSOGA & FA & FAGA & \cellcolor[HTML]{92D050}HFASSON & HPSOGA & FA & FA & FA & GWOCS \\
\textbf{2} & HFASSO & HFASSO & HFASSO & HFASSO & HFASSON & FA & GWOCS & FA & HFASSO & HFASSON & HFASSO & FAGA & FA & FAGA & FA & HPSOGA & HPSOGA & HPSOGA & HSSOGSA & GWOCS & GWOCS & FAGA & FA \\
\textbf{3} & SSO & SSO & SSO & GWOCS & SSO & FAGA & FA & GWOCS & SSO & SSO & SSO & HPSOGA & GWOCS & FA & SSO & GWOCS & HSSOGSA & GWOCS & FAGA & HSSOGSA & HSSOGSA & GWOCS & FAGA \\
\textbf{4} & GWOCS & GWOCS & HPSOGA & SSO & HFASSO & HPSOGA & FAGA & HPSOGA & FAGA & GWOCS & GWOCS & GWOCS & SSO & SSO & GWOCS & HSSOGSA & WCMFO & FA & WCMFO & FAGA & FAGA & SSO & HSSOGSA \\
\textbf{5} & FA & FA & GWOCS & FAGA & FA & GWOCS & HPSOGA & WCMFO & GWOCS & FA & FA & FA & HPSOGA & HSSOGSA & HFASSON & SSO & GWOCS & HSSOGSA & FA & HPSOGA & SSO & HSSOGSA & HFASSO \\
\textbf{6} & FAGA & HPSOGA & FA & HPSOGA & HPSOGA & SSO & HSSOGSA & HSSOGSA & HPSOGA & FAGA & HPSOGA & SSO & HFASSON & HFASSON & FAGA & FAGA & HFASSON & SSO & GWOCS & SSO & HPSOGA & HPSOGA & SSO \\
\textbf{7} & HPSOGA & FAGA & FAGA & FA & HSSOGSA & HFASSO & HFASSON & SSO & FA & HPSOGA & FAGA & HFASSO & HFASSO & HFASSO & HSSOGSA & HFASSON & FA & HFASSO & SSO & HFASSO & HFASSO & HFASSO & WCMFO \\
\textbf{8} & HSSOGSA & HSSOGSA & HSSOGSA & WCMFO & FAGA & HSSOGSA & HFASSO & HFASSO & HSSOGSA & HSSOGSA & HSSOGSA & HSSOGSA & HSSOGSA & WCMFO & HFASSO & HFASSO & HFASSO & FAGA & HFASSO & HFASSON & HFASSON & HFASSON & HFASSON \\
\textbf{9} & WCMFO & WCMFO & WCMFO & HSSOGSA & WCMFO & WCMFO & WCMFO & HFASSON & WCMFO & WCMFO & WCMFO & WCMFO & WCMFO & GWOCS & WCMFO & WCMFO & SSO & WCMFO & HFASSON & WCMFO & WCMFO & WCMFO & HPSOGA \\ \hline
\end{tabular}%
}
\end{table*}

\section{Application of the proposed Algorithm}
The performance of HFASSON has been shown to be superior with the CEC 2017 test suites and statistical tests; however, its effectiveness must be further evaluated on real-world problems. Therefore, the Cognitive Radio-Vehicular Ad Hoc Network (CR-VANET) algorithm is chosen to assess the performance of HFASSON. A comparison is made between the CR-VANET algorithm with cognitive radio spectrum sensing (CR-VANET) and the HFASSON applied to CR-VANET (CR-VANET\_HFASSON). The evaluation is performed using spectrum utilization of CR-VANET. 
\subsection{Background and problem formulation}
Cognitive radio applied to VANET (CR-VANET) for sensing vacant spectrum is an effective approach to reduce traffic and enabling real-time communication among vehicles in a VANET environment. Cognitive Radio (CR) manages the increasing demand for frequency spectrum usage by identifying unoccupied frequency bands for data transmission. In December 2003, the FCC recognizes cognitive radio as a candidate technology for implementing spectrum sharing \cite{Zeng2009}. The main functions of CR are: (i) spectrum sensing and analysis, and (ii) spectrum allocation and sharing\cite{Iverson2016}.
\begin{figure}[!ht]
\centering
\includegraphics[width=8 cm]{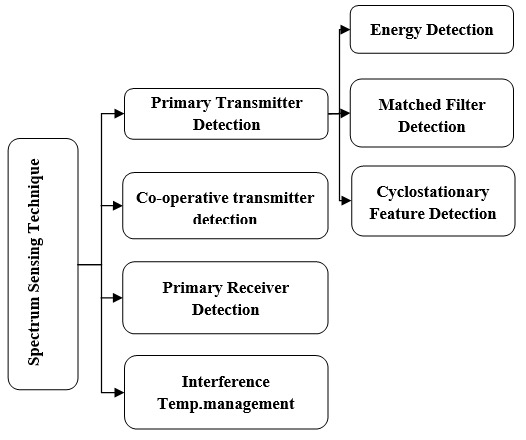}
\caption{ Spectrum sensing techniques\cite{10.1016/j.comnet.2006.05.001}.}
  \label{Spectrumsensing}
\end{figure}
Within the realm of cognitive radio (CR) operations, spectrum sensing is of significant importance. The main purpose of spectrum sensing is to enable a cognitive radio to sense information about its environment and the availability of vacant spectrum. Spectrum sensing technique is classified as primary transmitter detection, co-operative transmitter detection, primary receiver detection, and interference based sensing\cite{10.1016/j.comnet.2006.05.001}. Energy detection, Matched Filter detection and cyclostationary feature detection comes under primary transmitter detection method\cite{Bani2022}. Among the spectrum sensing techniques(Fig.\ref{Spectrumsensing}) matching filter detection and energy detection are frequently employed. 
\subsubsection{Energy Detection Method}
Energy detection (ED) \cite{Atapattu2014} \cite{Kockaya2020} is the widely used technique for spectrum sensing due to its simplicity and its independence from information about the primary signals. It is formulated based on a noise threshold. The spectrum occupancy in the ED process is determined by the threshold that is set. The energy detector makes its decision based on the following hypothesis test.
\begin{equation}
    H0: s(n) = G(n) 
\end{equation}
\begin{equation}
    H1: s(n)=P(n) + G(n) 
\end{equation}
Where
\begin{itemize}
    \item H0 represents the absence of primary user signal.
     \item H1 represents the presence of primary user signal.
\item s(n) is the signal received at each secondary user (SU). 
\item P(n) is the primary user (PU) signal.
\item G(n) denotes the additive white Gaussian noise (AWGN). 
\end{itemize}
In order to decide between the two states( H0 and H1 ) and determine the presence of primary user's signal, the threshold and perceived energy are compared to identify the primary user's presence. In energy detection, a bandpass filter with bandwidth “G(n)” prefiltered the received signal. Over a predefined time, the output signal of Analog to Digital Converter (ADC) is squared and integrated. An energy detection test statistic (Ted) is formulated using the resultant signal and is calculated as follows.
\begin{equation}
    Ted=\sum ^{N}_{n=0}\left| s\left( n\right) \right| ^{2}
\end{equation} 
Where N is the number of samples over a detection period.
The Probability of detection (Pd) represents the likelihood that the primary user's presence will be effectively detected, while the Probability of false alarm (Pfa) denotes the likelihood that the primary user's presence is falsely detected.
To ascertain the presence of a primary signal, the threshold ($\mu$) is compared with Ted. Pd and Pfa are determined using test statistics under the binary hypothesis to describe the energy detector's effectiveness. The evaluation of the effectiveness is determined by the parameters, Pd and Pfa, which are derived from test statistics within the framework of binary hypothesis.   
The Pd and Pfa is are given by, \cite{Verma2016}, 
\begin{equation}
    Pd=P[ Ted >  \mu / H_{1}]
\end{equation}
\begin{equation}
    Pfa=P[ Ted >  \mu / H_{0}]
\end{equation}
\begin{figure}[!ht]
\centering
\includegraphics[width=8.5cm]{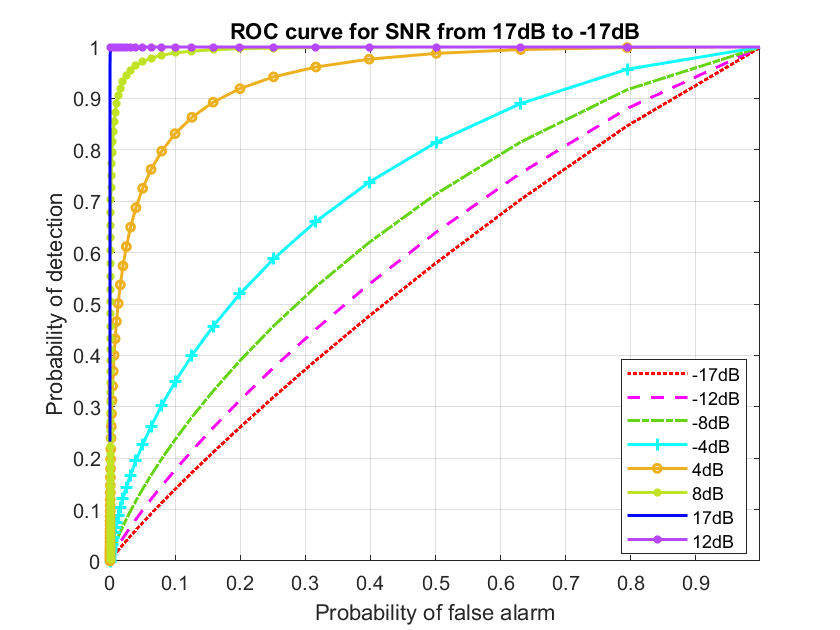}
\caption{ Variation of Pfa and Pd in ROC curve with range of SNR from 17dB to -17dB for spectrum sensing using energy detection model.}
  \label{ROC}
\end{figure}
Receiver operating characteristic (ROC) curve usually used to typify the performance of spectrum sensing.  ROC curve of Fig.\ref{ROC} shows the variation of Pd against Pfa during spectrum sensing using energy detection model for varying signal-to- noise ratio (SNR) that ranges from -17dB to 17dB. Each curve represents different values of the SNR. Moreover, as the value of SNR increases, the curve moves closer to the top left corner of the plot, which in turn specifies that Pd increases and Pfa decreases. The ideal case of the sensing algorithm has a value of 1 for Pd and 0 for Pfa, which is a straight line from 0 to 1 along the Y axis. In practice, it is impossible to reach an ideal value. Hence, there will be a trade-off between Pd and Pfa. 
\subsubsection{Objective function} 
Reducing the probability of incorrectly identifying spectrum holes is crucial for effective spectrum detection. Therefore, the main goal of the objective function in the spectrum sensing process is to minimize false alarms in order to enhance spectrum sensing performance. Consider the fitness value as'\( f_{\text{min}} \)'. The fitness function is formulated as follows: 
\begin{equation}
  f_{\text{min}}(j) =Thres(j)*-Pfa(j)*(\beta_{0}*\delta*\alpha), 0 < j \le N   
\end{equation}
Where 
\begin{itemize}
    \item 'Thres' represents a threshold value that one wants to achieve in the optimization problem. 
    \item 'Pfa' is the probability of false alarm. Shows the probability of incorrectly identifying a state or an occurrence.
    \item '$\beta_{0},\delta,\alpha$' are coefficients or constants that are used to establish the sensitivity of the fitness function and quantify the contributions of Pfa.
    \item N denotes the number of vehicles considered in the optimization problem.  
\end{itemize}
This formulation ensures that the fitness value is minimized while considering the false alarm probability and other coefficients that influence the optimization process. 
The presence of numerous applications in the vehicular network could lead to a degradation in communication efficiency and a scarcity of available spectrum. CR technology is now being applied to vehicular networks due to its ability to solve problems related to network capacity and exhaustion in unlicensed network frequency bands. As a result, CR-VANET facilitates effective vehicular communication and better spectrum utilization. The integration of CR in VANET makes use of spectrum holes to mitigate the effects of spectrum scarcity caused by improper allocation.
CR-VANET takes advantage of the use of the CR technique. Cognitive radio functions such as detecting spectrum opportunities, channel selection according to applications, and transmission of data without interfering with the primary user access. An efficient cooperative communication between vehicles is possible with a predefined road pattern in CR-VANET. In CR-VANET, the main challenge lies in the fact that spectrum sensing is primarily dependent on vehicle velocity, as opposed to conventional cognitive radio. 
\subsection{Result Analysis of CR-VANET\_HFASSON}
The performance metric considered for the result analysis of CR-VANET\_HFASSON is spectrum utilization. The output of this metric is compared with the CR-VANET algorithm. The evaluation of the metrics is performed for 20, 40, 60, 80, and 100.The parameter values used in CR-VANET\_HFASSON is given in Table \ref{table2}.
\subsubsection{Spectrum Utilization Vs Vehicle Count (VC) Analysis}
Spectrum utilization\cite{Verma2016} is the ratio of the sum of channel usage time by the vehicles of all channels to the channel maximum available time.
The formula for spectrum utilization (in percentage):
\begin{equation}
   \text{Spectrum Utilization} = \left( \frac{\text{Channel Usage}_{\text{Time}}}{\text{Total Time}_{\text{Available}}} \right) \times 100
\end{equation}
Where:
\begin{itemize}
    \item ${\text{Channel Usage}}_{\text{Time}}$ is the sum of channel 
    usage time for all channels
    \item $\text{Total Time}_{\text{Available}}$ is the product of the number of channels and the simulation time. 
\end{itemize}

\begin{table}[]
\caption{Comparison of average spectrum utilization (in \%) between CR-VANET and CR-VANET\_HFASSON.}
\label{spectrum utilization}
\begin{tabular}{lll}
 \hline
\textbf{Number of  Vehicles} & \textbf{CR-VANET} & \textbf{CR-VANET\_HFASSON} \\  \hline
20                           & 85.68\%           & \textbf{87.07\%}            \\
40                           & 89.2\%            & \textbf{90.18\%}         \\
60                           & 88.56\%           & \textbf{90.60\%}            \\
80                           & 90.48\%           & \textbf{91.18\%}            \\
100                          & 90.72\%           & \textbf{91.86\%}  \\  \hline       
\end{tabular}
\end{table}

The Table.\ref{spectrum utilization} summarizes  proportion of spectrum utilization of CR-VANET and CR-VANET\_HFASSON. Overall, the analysis of Table.\ref{spectrum utilization} shows that the spectrum utilization of CR-VANET\_HFASSON excels the other algorithm. CR-VANET\_HFASSON has 87.07\% spectrum utilization for 20 Vehicles which is 1.39\% higher than CR-VANET algorithm. For 40 vehicles, CR-VANET performs well and scores 0.72\% more than the CR-VANET\_HFASSON. However, the CR-VANET\_HFASSON  achieved spectrum utilization values of 2.04\%, 0.7\%, and 1.13\% for 60, 80, and 100 vehicles, respectively. These values are greater than those achieved by the CR-VANET.

\section{Conclusion} 
The proposed algorithm, HFASSON, combines the exploring skills of FA with the exploitation skills of SSO. Additionally, the suggested technique incorporates N-R method to enhance convergence potential while boosting its performance. The performance of HFASSON is compared with FA, SSO, HFASSO, WCMFO, HPSOGA, HSSOGSA, GWOCS, and FAGA using 23 benchmark functions in 30, 50, and 100 dimensions. 

The algorithms are compared using both qualitative and quantitative research techniques. The quantitative analysis employed best fitness metrics, while the qualitative study compared the HFASSON convergence rates with the above-mentioned eight algorithms . The experimental results show that the convergence speed of HFASSON is faster than that of HFASSO, FA, and SSO, as well as five other selected hybrid algorithms. 

The proposed method proved to be a viable alternative, as HFASSON outperformed eight for 30D and nine for 50 and 100D out of the 23 selected benchmark functions across broad and constrained search spaces. Applying the N-R method to refine the outcomes produced by the HFASSO algorithm further improved its accuracy and efficiency. Result of statistical test clearly indicates the outstanding performance of the proposed HFASSON algorithm. \\ \indent Moreover, the proposed algorithm proved to be better performing with the spectrum sensing technique in CR-VANET. Thus implementing CR-VANET with HFASSON (CR-VANET\_HFASSON), the simulation results indicate that CR-VANET\_HFASSON achieves enhanced spectrum utilization compared to the CR-VANET algorithm. \\ \indent The objective models used in this research may have certain limitations. In a real-world environment, additional factors could arise that might affect the results. Therefore, to validate the performance of HFASSON, we plan to implement CRVANET in a real environment using actual sensors and devices in the future. Furthermore, HFASSON can be expanded to incorporate hybrid optimization algorithms based on multiple objectives. \\ 

\section*{Acknowledgment}
This work is supported in part by the Malaysian Ministry of Higher
Education through the Fundamental Research Grant Scheme under Grant FRGS/1/2021/ICT11/UM/02/1(FP005-2021).
We would also like to express our gratitude to our university, Universiti Malaya, for their continued support in facilitating access to this grant.

\bibliographystyle{unsrt}  
\bibliography{export}

\end{document}